# A Replicable Robotics Awareness Method Using LLM-Enabled Robotics Interaction: Evidence from a Corporate Challenge

**S. A. Prieto [a], M. A. Gopee [b], Y. Ben Arab [a], B. García de Soto [b], J. Esteba [c] and P. Olivera Brizzio [c]**

[a] KINESIS Core Technology Platform, New York University Abu Dhabi (NYUAD), United Arab Emirates (UAE).
E-mail: samuel.prieto@nyu.edu, yb2300@nyu.edu
[b] S.M.A.R.T. Construction Research Group, Division of Engineering, New York University Abu Dhabi (NYUAD), United Arab Emirates (UAE).
E-mail: amg1289@nyu.edu, garcia.de.soto@nyu.edu
[c] AD Ports Group, Corporate Innovation Department, United Arab Emirates (UAE).
E-mail: joan.esteba@adports.ae, pablo.brizzio@adports.ae

## Abstract

Large language models are increasingly being explored as interfaces between humans and robotic systems, yet there remains limited evidence on how such technologies can be used not only for interaction, but also as a structured means of introducing robotics to non-specialist users in real organizational settings. This paper introduces and evaluates a challenge-based method for robotics awareness, implemented through an LLM-enabled humanoid robot activity conducted with employees of AD Ports Group in the United Arab Emirates. In the event, participants engaged with a humanoid robot in a logistics-inspired task environment using voice commands interpreted through an LLM-based control framework. The activity was designed as a team-based, role-driven experience intended to expose participants to embodied AI and human-robot collaboration without requiring prior robotics expertise. To evaluate the approach, a post-event survey remained open for 16 days and collected 102 responses. Results indicate strong overall reception, with high satisfaction (8.46/10), increased interest in robotics and AI (4.47/5), and improved understanding of emerging forms of human-robot collaboration (4.45/5). Participants who interacted directly with the robot also reported natural interaction (4.37/5) and a strong sense that interaction became easier as the activity progressed (4.74/5). At the same time, lower ratings for reliability and predictability point to important technical and design challenges for future iterations. The findings suggest that challenge-based, LLM-enabled humanoid interaction can serve as a promising and replicable method for robotics awareness in industrial and operational environments.

## 1. Introduction

Recent advances in large language models (LLMs) have opened new possibilities for human–robot interaction by enabling users to communicate with robotic systems through natural language. Instead of relying on structured programming, dedicated interfaces, or expert supervision, LLM-enabled systems can interpret conversational instructions and translate them into executable robot actions. This capability has the potential to reduce technical barriers to robot use and to make robotic platforms more accessible to users with little or no prior robotics expertise.

Several recent research efforts have explored the integration of LLMs into robotic control architectures. For example, systems such as SayCan combine large language models with robot skill libraries to select feasible actions that satisfy user-specified goals expressed in natural language (Ahn et al., 2022). Similarly, embodied multimodal models such as PaLM-E integrate language reasoning with visual perception and robotic control, enabling robots to interpret environmental context while executing tasks (Driess et al., 2023). More recent work has extended this paradigm through the development of vision–language–action models capable of directly generating robot actions from natural language instructions. For instance, the RT-2 model demonstrates how knowledge learned from large-scale web data can be transferred to real-world robotic manipulation tasks (Brohan et al., 2023). Other approaches focus on grounding spatial reasoning and manipulation planning through language-based representations. VoxPoser, for example, introduces composable three-dimensional value maps that allow robots to plan manipulation tasks guided by natural language instructions (Huang et al., 2023). Together, these developments suggest that natural language is becoming a viable interface layer between humans and increasingly capable robotic systems.

At the same time, progress in robot intelligence does not automatically translate into broader organizational understanding, acceptance, or readiness for adoption. In many operational environments, including logistics, ports, construction, and industrial settings, the challenge is not only whether a robot can perform a task, but also whether non-specialist users can meaningfully understand its capabilities, limitations, and potential roles in real workflows. This is especially important as robots move beyond research laboratories and become more visible in shared professional environments. Prior work has emphasized the importance of intuitive interaction mechanisms and clear human-robot relationships in collaborative settings, including broader perspectives such as human-robot partnership (Prieto et al., 2024; Yu et al., 2024). In parallel, research on industrial human-robot collaboration highlights that successful integration depends on human-centered design, worker involvement, and gradual familiarization with robotic systems (Krüger et al., 2009). Industry-led deployments further reinforce this perspective. For example, organizations such as Amazon have introduced large-scale mobile robotic systems in warehouse environments where robots and human workers operate in tandem, with robots transporting inventory to human operators and enabling new forms of collaborative workflows (Pasparakis et al., 2023; Wurhofer et al., 2015). In these settings, robots are typically designed to handle repetitive or physically demanding tasks, allowing workers to focus on more complex activities while progressively adapting to robot-supported processes. Similarly, manufacturers such as BMW and Ford Motor Company have explored the use of collaborative robots (cobots) in assembly and production lines, where robots are deployed alongside human workers in shared workspaces to support cooperative task execution and gradual workforce familiarization with automation technologies (Krüger et al., 2009). These examples illustrate how real-world deployments increasingly rely on incremental, interaction-based exposure to robotics, where understanding emerges through day-to-day collaboration rather than through isolated training. From a broader organizational perspective, studies on Industry 4.0 adoption similarly identify experiential exposure and technology familiarization as key factors in reducing resistance and supporting informed adoption (Sony & Naik, 2020). Despite these developments, such exposure is often informal, unstructured, or embedded in deployment processes, and therefore does not necessarily provide a replicable framework for systematically building robotics awareness among non-specialist users.

Challenge-based and event-driven formats provide an additional relevant perspective. Robotics competitions such as the DARPA Robotics Challenge and RoboCup have demonstrated the value of task-driven engagement for stimulating innovation and supporting knowledge dissemination (Kitano et al., 1998). Related work suggests that goal-oriented activities can help participants develop concrete mental models of robotic behavior through active involvement, while studies in human–robot interaction show that even short-term hands-on experiences can significantly influence perceptions of trust, usability, and willingness to engage with robots (Hancock et al., 2011; Pedersen et al., 2016). However, much of this literature focuses on research communities, educational settings, or public demonstrations rather than on non-specialist personnel operating within real organizational environments. As a result, there remains limited work on replicable methods that combine structured participation, direct interaction, and operationally meaningful tasks to introduce robotics to non-expert users in workplace contexts.

This gap is particularly relevant for humanoid robots and other embodied AI systems. Their physical form and natural-language interfaces create strong opportunities for accessibility and engagement, but they also raise practical questions about trust, predictability, usability, and perceived value in real organizational contexts. Existing studies have begun to examine LLM-enabled interaction in applied robotics domains, including speech-driven perception, navigation, and safety-oriented applications (Gopee et al., 2025, 2026). Yet much of the literature remains focused on technical performance, system architecture, or task execution. Comparatively less attention has been given to the design of replicable methods that allow organizations to expose non-expert personnel to embodied AI in ways that can stimulate awareness, reflection, and early identification of application opportunities. In this sense, natural-language interaction may serve not only as a control mechanism, but also as an accessibility layer that enables more intuitive and inclusive forms of engagement with robotic systems.

To address this need, this paper introduces and evaluates a challenge-based approach for robotics awareness implemented through an LLM-enabled humanoid robot activity conducted with employees of AD Ports Group and its Corporate Innovation Department in the United Arab Emirates. The challenge was designed as a team-based, logistics-inspired interaction experience in which participants guided a humanoid robot through navigation and manipulation tasks using voice commands, without requiring programming knowledge or prior hands-on experience with robotics. Rather than treating the activity solely as a demonstration, the approach was structured as an interactive method to expose participants to the practical realities of human–robot collaboration, including communication, coordination, task execution, and system limitations.

The paper evaluates this approach through a post-event survey that captured perceptions from 102 respondents across attendees and direct participants. In addition to assessing user experience during interaction, the study examines whether the challenge increased interest in robotics and artificial intelligence, improved understanding of emerging forms of human–robot collaboration, and revealed design factors that may influence future adoption and deployment. In this way, this paper contributes not only empirical observations about LLM-enabled humanoid interaction, but also a replicable framing for robotics awareness activities in operational and industrial environments.

## 2. Proposed Robotics Awareness Method and System Implementation

This section presents the framework used in the Humanoid Challenge, combining a structured robotics awareness method with a practical implementation based on LLM-enabled humanoid interaction. The objective was not only to create an engaging challenge scenario, but also to provide participants with an accessible and operationally meaningful introduction to embodied AI and human-robot collaboration. To clarify both the conceptual and technical dimensions of the approach, the section first introduces the methodological rationale, then describes the system implementation, and finally details the challenge environment and task structure.

### 2.1. Robotics Awareness Method

The activity was designed not only as a robot demonstration, but as a structured robotics awareness method intended to expose non-specialist participants to embodied AI and human-robot collaboration in an operationally meaningful setting. Rather than observing the robot passively, participants were required to engage directly with the system through spoken instructions, coordinate as a team, and adapt their communication based on the robot's responses and task progress. This allowed participants to experience both the accessibility and the practical constraints of natural-language robot control.

The method was guided by four design principles. First, interaction had to be accessible to users without programming or robotics expertise. Second, the scenario had to be operationally recognizable, drawing on simplified logistics and port-related activities rather than abstract laboratory exercises. Third, participation had to be distributed across multiple team members in order to encourage shared engagement and reflect the collaborative nature of real work environments. Fourth, the activity had to be observable and evaluable, so that both participant reflection and post-event assessment could be supported.

In this paper, robotics awareness is understood as increased familiarity with robotic capabilities, limitations, interaction requirements, and potential application opportunities among users who are not necessarily robotics specialists. The challenge format was used to promote this awareness through active exposure rather than passive observation. By combining direct interaction, time-constrained decision-making, task-oriented communication, and visible robot behavior within a simplified but recognizable operational scenario, the activity encouraged participants to form a more concrete understanding of what current robotic systems can and cannot do in practice.

This framing is important because the value of the challenge extended beyond the immediate task itself. The activity was intended to act as a bridge between emerging robotic technologies and potential end users in an industrial setting, offering a controlled but realistic environment in which participants could engage with a humanoid robot, reflect on its usability, and begin to identify possible applications in their own operational contexts.

### 2.2. System Implementation

Participants interacted with a Unitree G1 humanoid robot (Figure 1) integrated within a modular, LLM-enabled control architecture that mapped natural language input to executable robot actions. The system followed a multi-stage pipeline consisting of automatic speech recognition (ASR), language understanding via a large language model (LLM), structured command generation, and low-level robot execution through the manufacturer's software development kit (SDK).

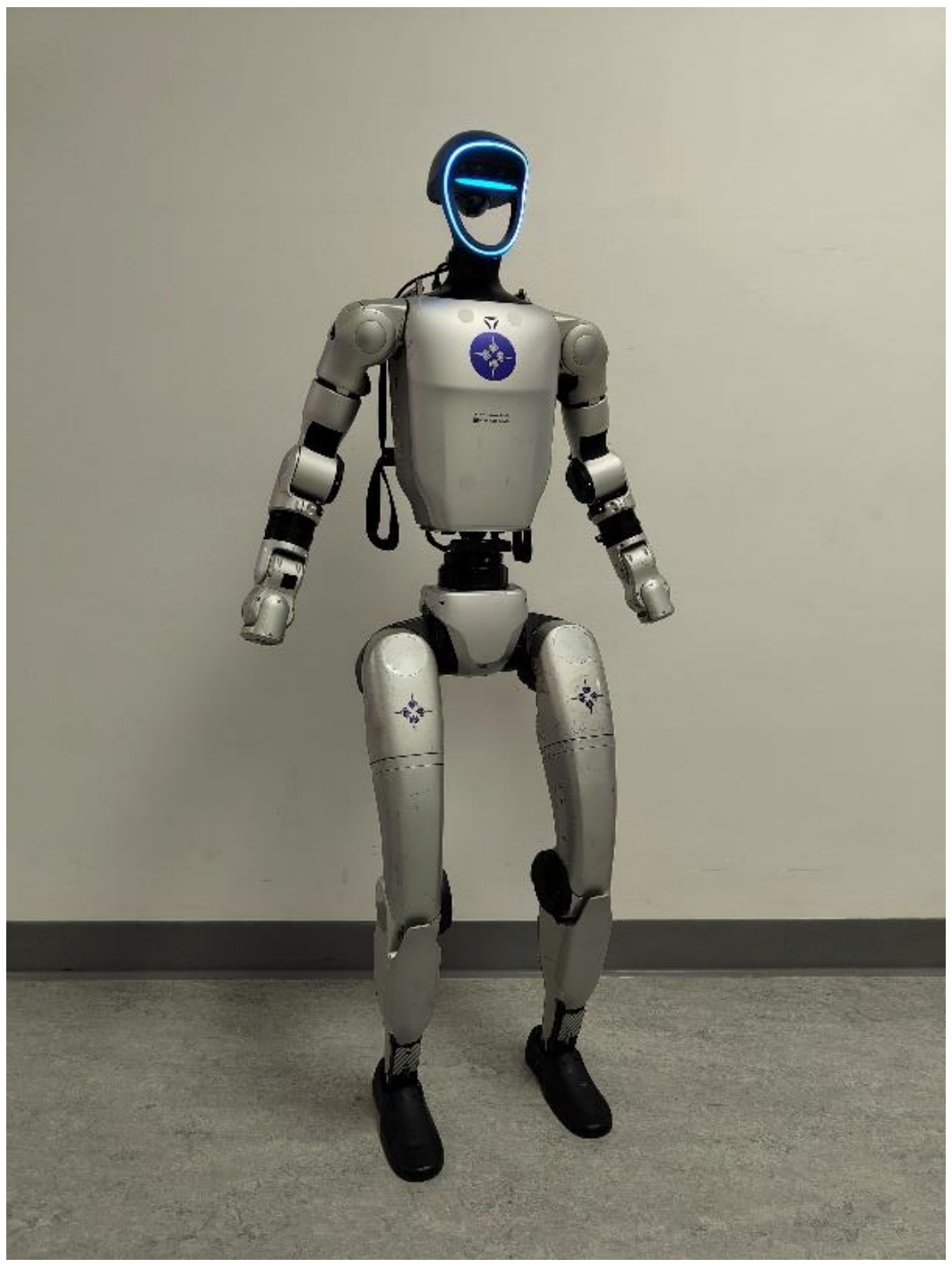

*Figure 1. Unitree G1 humanoid robot used for the challenge*

Speech input was captured using a push-to-talk interface and processed by an ASR engine to produce a text transcription. This transcription was then passed to the LLM, which functioned as a semantic parser and reasoning module. The LLM was prompted using a constrained instruction format that defined the available action space (e.g., navigation, rotation, grasping) and expected output schema. Specifically, the model generated structured intermediate representations (e.g., JSON-like command objects) encoding intent, action type, parameters (distance, angle), and optional contextual constraints. This approach reduced ambiguity and improved robustness by enforcing consistency between natural language input and executable commands

As shown in Figure 2, the system architecture is organized into a sequential processing pipeline that connects human input to robot execution through distinct functional modules. Interaction begins with the speech capture component, where the push-to-talk interface regulates when audio is recorded and forwarded to the ASR engine. The ASR module converts the audio signal into text, which is then processed by the LLM-based intent interpretation layer. At this stage, the LLM performs semantic parsing and maps the transcribed instruction into a structured command representation aligned with the predefined action schema. The resulting command is passed to the validation and control logic module, which enforces syntactic correctness and operational feasibility before forwarding validated commands to the robot SDK communication layer. The SDK interface then translates high-level commands into low-level control primitives, including motion commands and actuator signals, enabling execution across the robot's locomotion and manipulation subsystems. This modular decomposition highlights a clear separation between probabilistic components (ASR and LLM) and deterministic control layers, improving system robustness and interpretability.

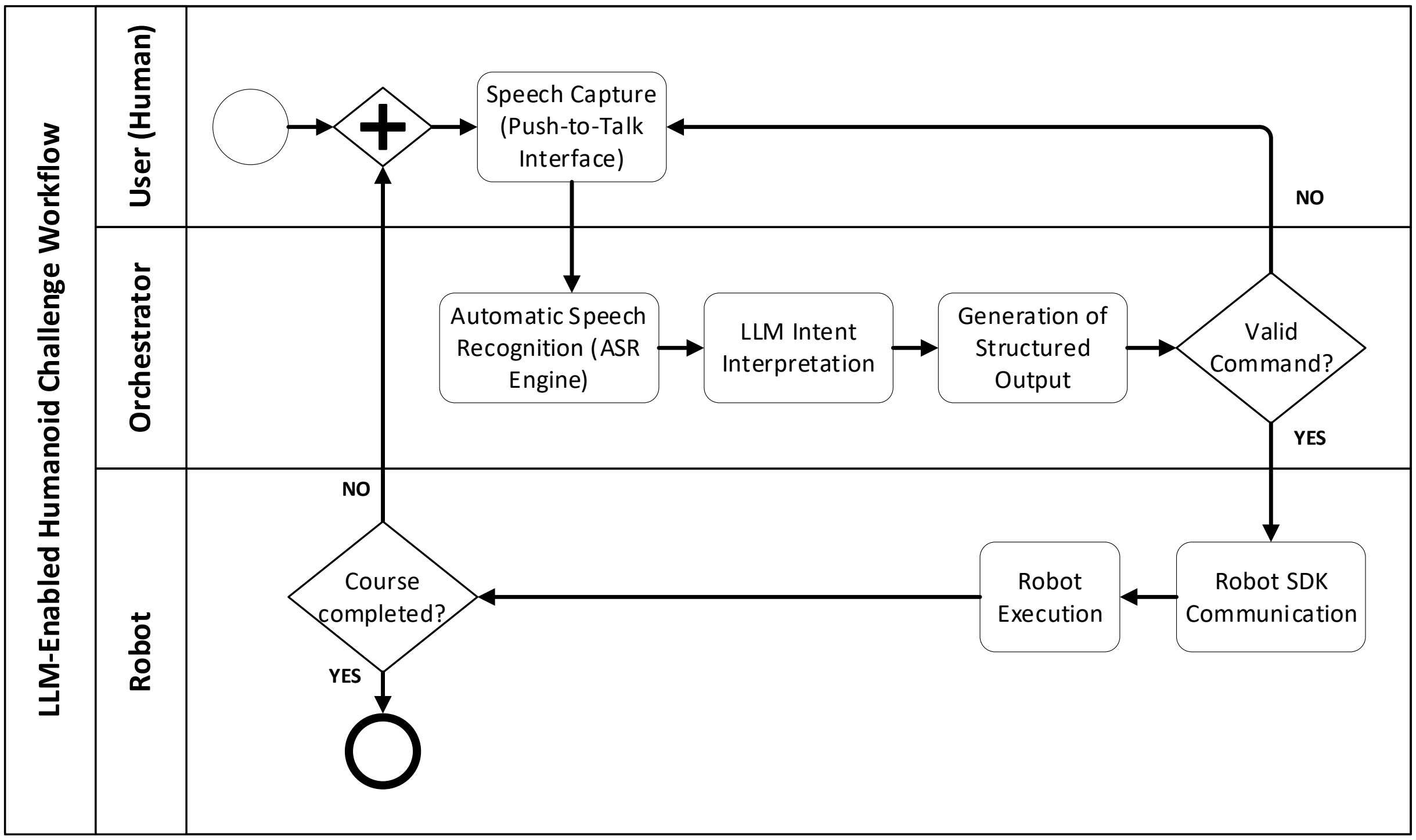

*Figure 2. BPMN diagram showcasing how communication happens between the human and the robot*

To ensure safe and reliable operation, a rule-based validation and control layer was implemented downstream of the LLM. This layer performed syntactic and semantic checks on generated commands, including parameter bounds, action feasibility, and conflict resolution (e.g., rejecting simultaneous incompatible actions). Only validated commands were forwarded to the robot SDK, where they were translated into low-level motion primitives and actuator controls. This separation between high-level reasoning and deterministic execution aligns with emerging architectures in LLM-enabled robotics that leverage structured intermediate representations to translate natural language into executable actions while maintaining safety and control constraints (Liang et al., 2022).

The robot supported a predefined but extensible command set spanning locomotion (forward/backward movement, lateral shifts, rotational control) and basic manipulation (grasp/release via an end-effector). Parameterized commands allowed users to specify quantitative values (e.g., "move forward 1 meter" or "turn 90 degrees"), which were normalized into the robot's coordinate and control system. Temporal sequencing was handled implicitly, with each validated command executed atomically before accepting the next instruction, thereby simplifying interaction for non-expert users.

To maintain real-time responsiveness, the system employed lightweight prompt engineering and constrained decoding strategies, minimizing latency in LLM inference. In cases where deployment-specific implementation details cannot be disclosed (e.g., model hosting configuration, fine-tuning procedures), the system design follows patterns consistent with prior work on on-device and hybrid LLM-based robotic control pipelines (Gopee et al., 2026; Liang et al., 2022).

Overall, the architecture operationalized a layered interaction paradigm (i.e., speech capture, semantic interpretation, structured command generation, validation, and execution) enabling intuitive yet controlled human-robot interaction suitable for real-time task environments (Figure 2). This design balances flexibility in natural language input with the determinism required for safe and predictable robot behavior.

### 2.3. Challenge Environment and Task Structure

The challenge environment was designed to simulate simplified logistics operations within a controlled indoor setting inspired by port and warehouse workflows. Within this environment (Figure 3), participants guided the Unitree G1 humanoid robot through a structured course involving navigation, object manipulation, obstacle negotiation, and delivery tasks. The setting was intentionally simplified so that participants could focus on interaction and coordination without requiring prior technical expertise, while still engaging with a scenario that felt operationally relevant.

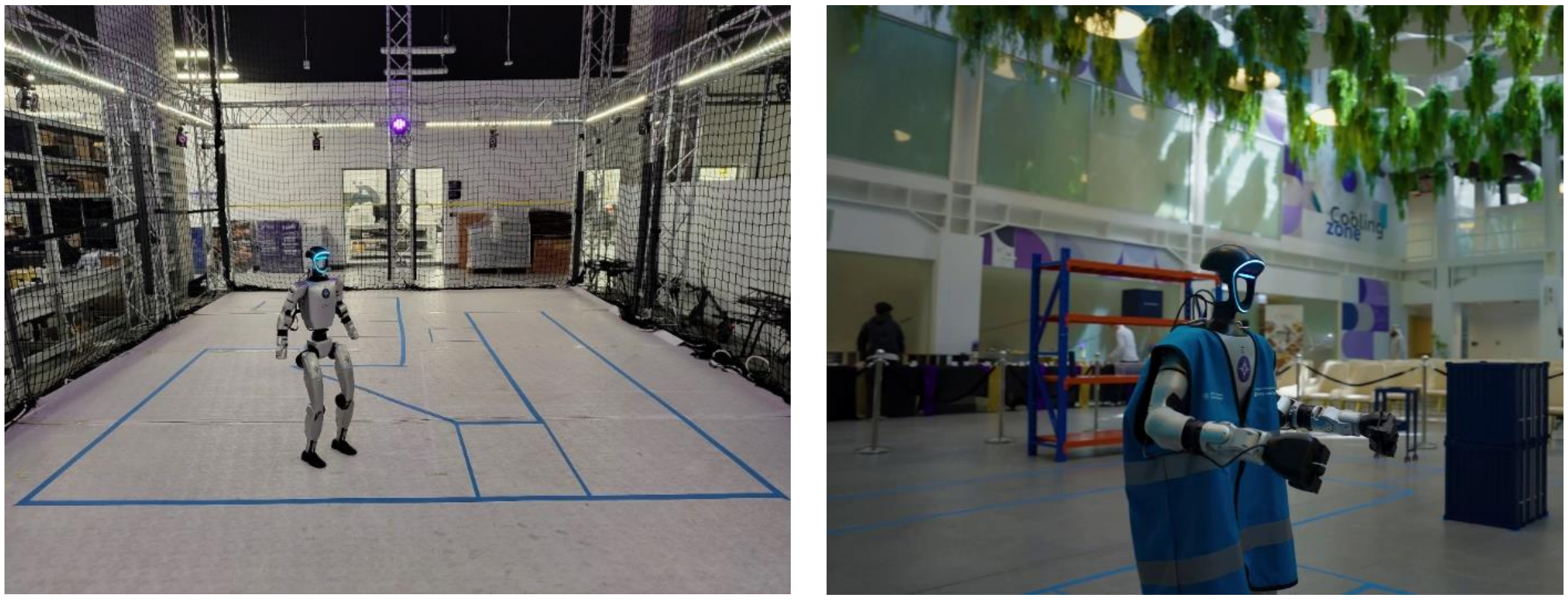

a) b)

*Figure 3. G1 robot within the a) test environment used for development and b) the final deployment at AD Ports Group installations during the competition day*

Each team consisted of five participants, and the challenge followed a relay-style format in which each team member controlled the robot during a specific stage of the course. This structure ensured broad participation while also reflecting the distributed nature of real operational workflows, in which different actors may contribute to different parts of a robot-supported process. The relay format also introduced a collaborative and social dimension that reinforced the broader awareness objective of the activity.

Each team was allocated a total of thirty minutes to complete the challenge. This period included five minutes for setup and operator positioning, ten minutes for familiarization and command testing outside the course area, and fifteen minutes of official run time during which the team attempted to complete the full course. The familiarization period was especially important, as it gave participants an opportunity to develop an initial understanding of the interaction style before entering the scored portion of the activity.

The course consisted of five sequential stages with 4 different checkpoints (Figure 4). In the first stage, the robot navigated from the starting position to the first checkpoint using basic locomotion commands. In the second stage (i.e., in between checkpoint 1 and 2), the robot was instructed to grasp a token provided by the operator. The third stage introduced obstacle navigation, requiring participants to guide the robot around an obstacle while maintaining controlled movement. This presented the participants with the choice of either risking the shortest path overcoming the object (risking hitting the object due to a narrower path), or going around the object in a wider, safer path, but taking more time doing so. In the fourth stage, the robot was required handle the token collected in stage 2, and then to pick up a box containing the token, given by one of the operators. In the fifth and final stage, the robot transported the box to a designated delivery area and handed it to the last operator while reaching the finish line.

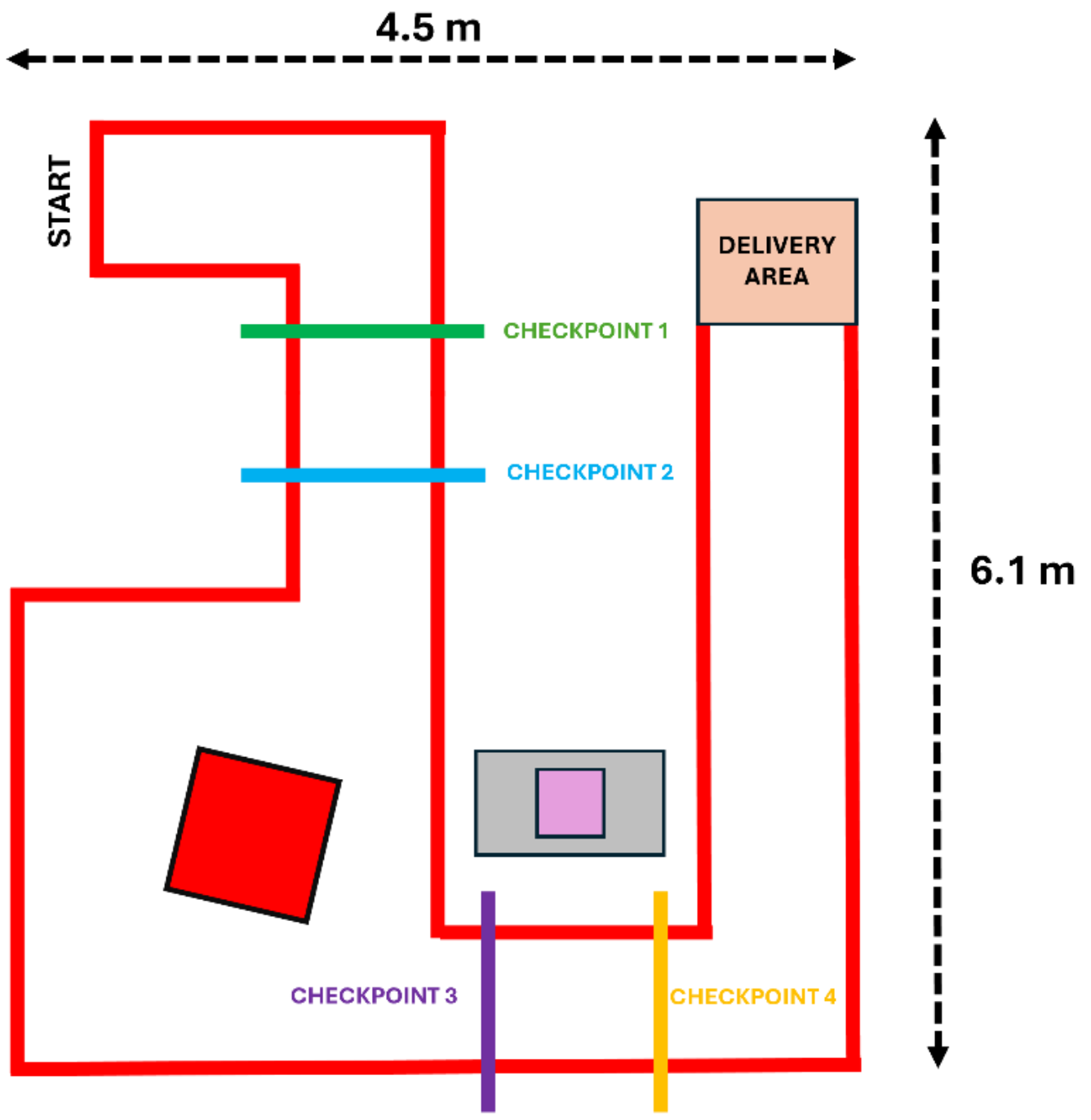

*Figure 4. Diagram of the circuit used during the challenge.*

Performance was evaluated through a time-based scoring system that combined total run time with penalties for operational errors. These penalties included events such as crossing navigation boundaries, colliding with obstacles, or requesting operator assistance. This scoring approach made it possible to account for both efficiency and execution quality, while also increasing the challenge value of the activity for both participants and observers.

Taken together, the methodological framing, the system implementation, and the challenge structure formed a replicable framework for exposing non-specialist participants to humanoid robotics in an interactive and operationally grounded way. The technical architecture enabled direct natural-language control, while the challenge design ensured that interaction occurred in a structured, collaborative, and evaluable format aligned with the paper's broader robotics awareness objective.

## 3. Evaluation Design and Data Collection

This section describes the evaluation approach used to assess the proposed robotics awareness method following the Humanoid Challenge. Because the purpose of the study was not limited to technical system performance, the evaluation focused on participant perceptions of the activity, the human-robot interaction experience, and the broader awareness outcomes generated by the event. In particular, the evaluation was intended to examine whether the challenge format was engaging and understandable, whether the LLM-enabled interaction was perceived as accessible and intuitive, and whether the activity contributed to increased interest in robotics and improved understanding of human-robot collaboration in operational contexts.

### 3.1. Survey Design

Following completion of the event, attendees were invited to complete a post-event questionnaire designed to capture both event-level impressions and perceptions of the humanoid robot interaction experience. The survey remained open for sixteen days and received a total of 102 responses. The questionnaire collected demographic information, professional background, and prior experience with robotics and voice-based technologies. It also included items related to the event itself, the interaction with the robot, and possible directions for future improvement.

Most close-ended items were measured using five-point Likert scales, allowing respondents to indicate the degree to which they agreed with statements related to accessibility, trust, safety, interaction quality, and interest in future engagement. Overall satisfaction with the event was measured separately using a ten-point rating scale. In addition to the quantitative items, the survey included open-ended questions that allowed respondents to identify positive aspects of the experience and suggest improvements for future iterations of the challenge. These qualitative responses were included to provide context for the numerical results and to support the identification of design implications for subsequent versions of the activity.

The structure of the questionnaire reflected the dual purpose of the study. On one hand, it aimed to assess the immediate user experience associated with natural-language interaction with the humanoid robot. On the other hand, it was intended to evaluate the broader effectiveness of the challenge as a robotics awareness activity for non-specialist participants. For this reason, the instrument included both interaction-focused items, such as perceived naturalness, control, transparency, predictability, trust, and mental effort, and event-level items related to interest in robotics and artificial intelligence, likelihood of future participation, and perceived understanding of emerging forms of human-robot collaboration.

### 3.2. Respondent Profile

The survey captured responses from a diverse group of participants and attendees associated with the event. Respondents represented a broad range of age groups, with the largest group being 25 to 34 years old, followed by 35 to 44 and 18 to 24. A smaller number of respondents fell within the 45 to 54 age group, and no respondents reported being above 55 years of age. In terms of gender distribution, 79 respondents identified as men and 23 as women. Respondents also reported varied professional and academic backgrounds, including robotics and artificial intelligence, engineering and computer science, human factors and human-computer interaction, education, and industry practice. This diversity is relevant because it reflects the interdisciplinary and mixed-expertise audience typically encountered in robotics demonstrations and applied human-robot interaction activities.

Prior experience with robotics among respondents was generally limited. Forty-one respondents indicated no prior experience with robotics, while 48 reported minimal experience, 11 reported moderate experience, and only 2 reported extensive experience. Familiarity with voice assistants was somewhat higher, with 22 respondents reporting no experience, 42 minimal experience, 29 moderate experience, and 9 extensive experience. These distributions suggest that the evaluation largely reflects the views of individuals with limited exposure to robotics, which is consistent with the paper's focus on awareness-building among non-specialist or lightly experienced users. At the same time, the somewhat higher familiarity with voice-based technologies may have reduced the novelty of the speech interface itself, allowing participants to focus more on the embodied interaction with the robot.

## 4. Results

A total of 49 responses were collected from attendants to the Humanoid Challenge. Within that group, 19 participated directly in the challenge and interacted with the robot, while 30 attended as observers. The results are therefore presented at two levels: event-level outcomes and direct interaction outcomes.

Event-Level Outcomes

The event was rated very positively across all event-level measures. The strongest outcomes were willingness to recommend the event, increased interest in robotics and AI, willingness to attend similar events again, and improved understanding of human-robot collaboration. These results support the interpretation of the challenge as an effective robotics awareness activity, not only as a one-time demonstration. A summary of the results can be seen in Figure 5 and Table 1.

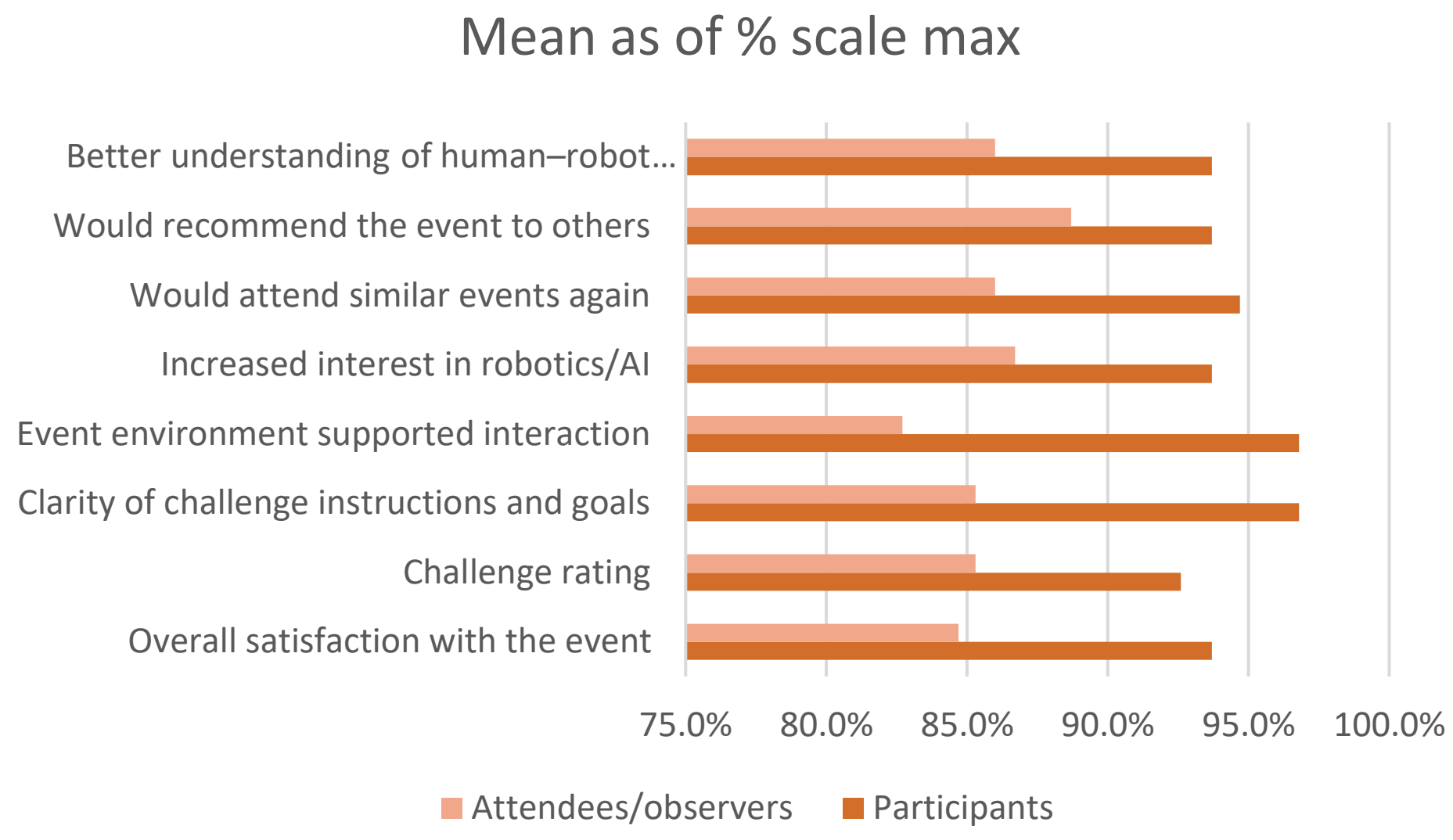

*Figure 5. Bar chart showcasing a summary of event-level outcomes split by attendee role (direct participants vs observers).*

*Table 1. Event-level survey results for the humanoid challenge. Columns indicate the response scale, mean score, standard deviation (SD) and number of respondents (N) split by attendee role (direct participants vs observers).*

| | | Direct Participants | | Observers | |
|---|---|---|---|---|---|
| **Item** | **Scale** | **Mean** | **SD** | **Mean** | **SD** |
| Overall satisfaction with the event | 10 | 9.37 | 1.67 | 8.47 | 1.36 |
| How do you rate the Humanoid Challenge | 5 | 4.63 | 0.96 | 4.27 | 0.69 |
| Instructions and goals of the challenge were clear | 5 | 4.84 | 0.37 | 4.27 | 0.64 |
| The event environment supported interaction well | 5 | 4.84 | 0.37 | 4.13 | 0.94 |
| The event increased my interest in robotics/AI | 5 | 4.68 | 0.95 | 4.33 | 0.61 |
| I would attend similar events again | 5 | 4.74 | 0.93 | 4.30 | 0.60 |
| Likelihood of recommending the event to others | 5 | 4.68 | 0.95 | 4.43 | 0.57 |
| The challenge helped me better understand the future of human–robot collaboration | 5 | 4.68 | 0.95 | 4.30 | 0.65 |

Direct Interaction Experience

Table 2 summarizes the direct interaction metrics reported by respondents who engaged hands-on with the robot. Figure 6 presents the same variables graphically, including perceived naturalness, robot understanding, sense of control, transparency, and predictability.

*Table 2. Direct interaction experience metrics reported by participants who interacted with the robot. Columns indicate the response scale, mean score, standard deviation (SD) and number of respondents (N).*

| **Item** | **Scale** | **Mean** | **SD** | **N** |
|---|---|---|---|---|
| The interaction with the robot felt natural | 5 | 4.37 | 0.96 | 19 |
| The robot understood my instructions well | 5 | 4.32 | 0.95 | 19 |
| I felt in control of the robot | 5 | 4.21 | 1.03 | 19 |
| The robot made it easy to understand what it was doing and why, transparency | 5 | 4.26 | 0.93 | 19 |
| The robot’s behavior was predictable | 5 | 4.00 | 1.00 | 19 |

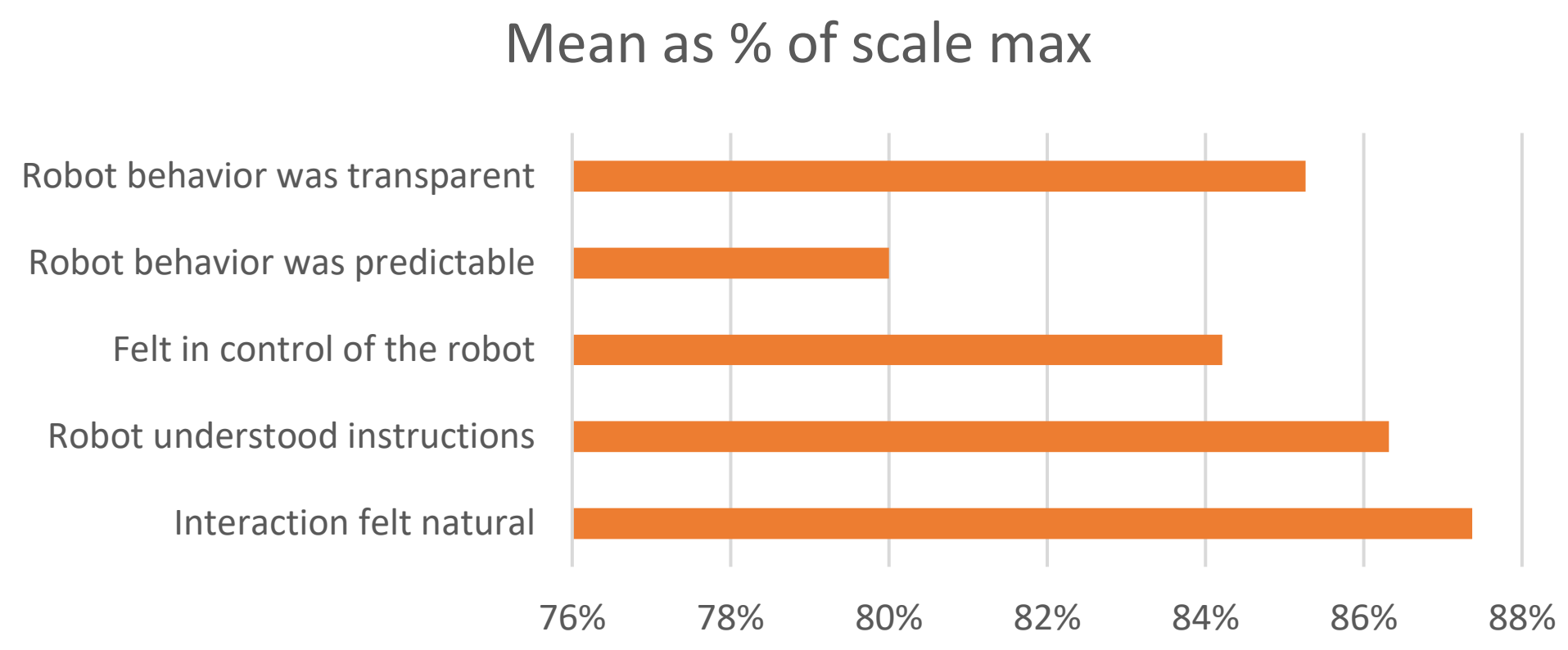

*Figure 6. Bar chart showing a summary of the direct interaction experience.*

Safety, Trust, and Interaction learning

As reported in Table 3. Safety, trust, cognitive effort, and learning metrics for direct participants. Columns indicate the response scale, mean score, standard deviation (SD) and number of respondents (N).Table 3 and Figure 7, physical safety during the interaction received a mean rating of 4.47, and comfort interacting with the robot again in real environments received a mean of 4.32. Trust in the robot to behave safely near people and objects received a mean of 3.95. The statement that the interaction required high mental effort received a mean of 3.53, while the statement that participants had to think carefully about how to phrase commands received a mean of 4.00. The statement that the interaction became easier as it progressed received a mean of 4.74.

*Table 3. Safety, trust, cognitive effort, and learning metrics for direct participants. Columns indicate the response scale, mean score, standard deviation (SD) and number of respondents (N).*

| Item | Scale | Mean | SD | N |
|---|---|---|---|---|
| I felt physically safe during the interaction | 5 | 4.47 | 1.02 | 19 |
| I would feel comfortable interacting with this robot again in real environments | 5 | 4.32 | 1.00 | 19 |
| I trusted the robot to behave safely near people and objects | 5 | 3.95 | 1.08 | 19 |
| The interaction required high mental effort | 5 | 3.53 | 1.31 | 19 |
| I had to think carefully about how to phrase commands | 5 | 4.00 | 1.15 | 19 |
| The interaction became easier as it progressed | 5 | 4.74 | 0.45 | 19 |

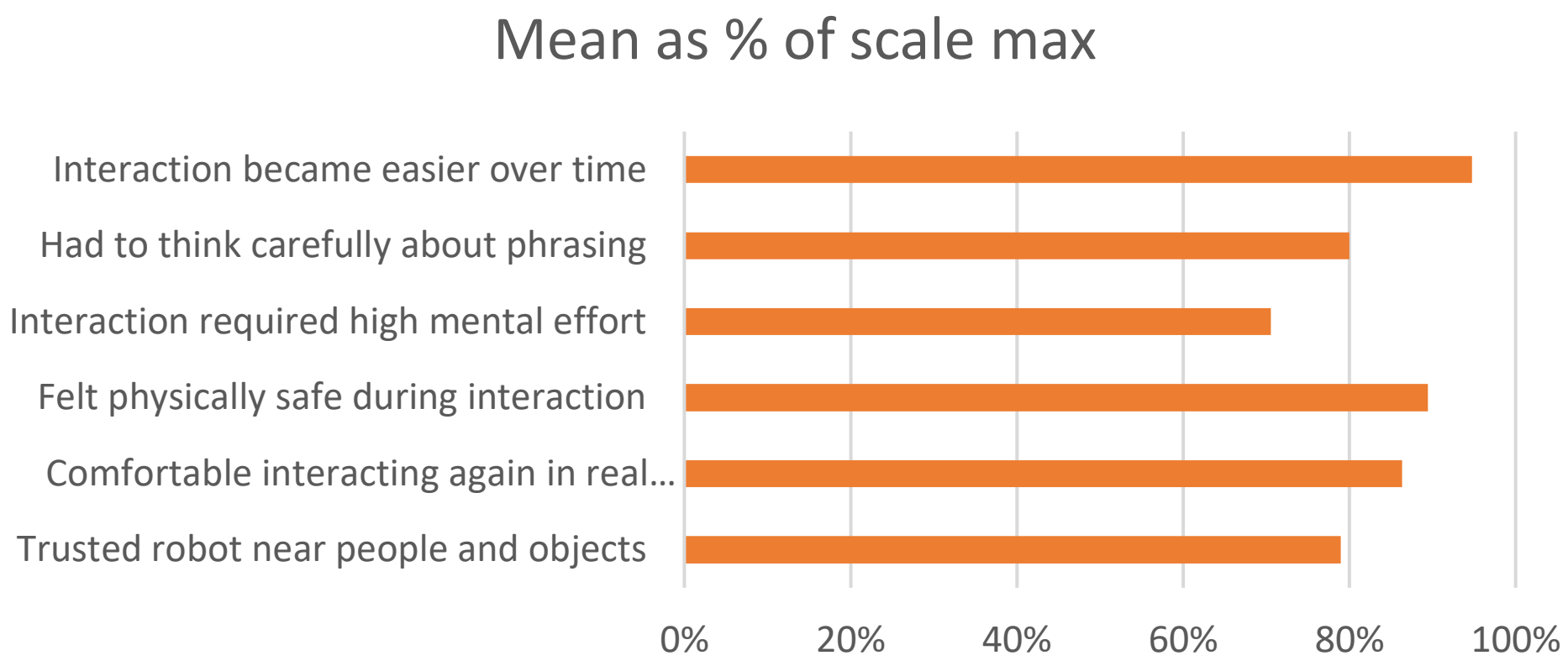

*Figure 7. Bar chart showing a summary of the safety, trust, and interaction learning results.*

<u>Qualitative Feedback</u>

The qualitative responses are summarized in Table 4 and Figure 8, which present the most frequently reported concerns.

*Table 4. Most frequently reported concerns related to interaction with the humanoid robot.*

| Concern theme | Count |
|---|---|
| Misinterpretation of commands | 11 |
| Reliability / failure risk | 10 |
| Over-trust in automation | 10 |
| Physical safety | 5 |
| Job displacement | 4 |
| Privacy | 3 |
| Lack of transparency | 2 |
| No concerns | 2 |

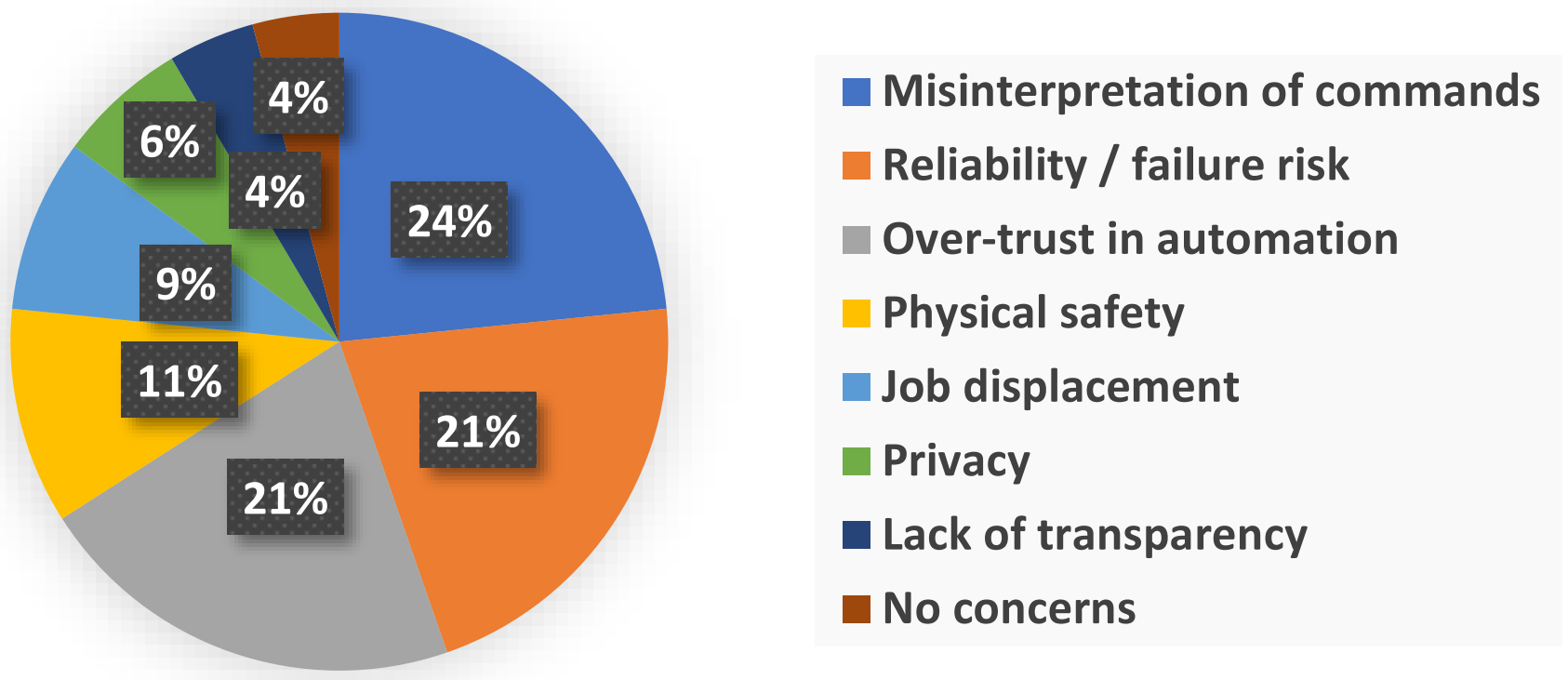

*Figure 8. Pie chart showing a summary of the qualitative feedback.*

## 5. Discussion

The overall picture that emerges from the results is that the Humanoid Challenge worked well as a first-contact experience with embodied AI for a largely non-specialist audience. The event-level outcomes were consistently strong, with high ratings for satisfaction, willingness to recommend the event, willingness to attend similar activities again, and increased interest in robotics and artificial intelligence, as shown in Table 1 and Figure 5. These results matter because the contribution of the activity does not lie only in the robot itself, but in its ability to create a structured setting in which participants could engage with robotics in a direct and memorable way.

That broader framing becomes even more relevant when considered alongside the respondent profile. Most respondents reported little or no prior experience with robotics, yet those who interacted directly with the system still rated the interaction as natural and reported that the robot generally understood their instructions. This supports the idea that natural-language interaction can lower the initial barrier to engagement, especially in organizational settings where potential users are unlikely to have technical robotics training. Rather than needing to learn a formal control interface, participants could begin from speech, which is already familiar to them as a communication modality.

This interpretation is consistent with technology-acceptance research, which identifies perceived ease of use, perceived external control, and anxiety as important factors in early engagement with new systems (Venkatesh & Bala, 2008). In that sense, allowing participants to interact through speech rather than through a formal control interface likely reduced some of the initial friction associated with an unfamiliar robotic system. At the same time, research on automation shows that trust, workload, and perceived reliability strongly shape how people engage with automated systems (Parasuraman & Riley, 1997). Although the present study did not measure long-term adoption directly, the challenge format can therefore be understood as a useful first exposure to both the accessibility and the constraints of LLM-enabled robot interaction.

At the same time, the results do not suggest that natural language makes interaction effortless. Participants still reported moderate mental effort and noted that they had to think carefully about how to phrase commands. The combination of these items with the strong rating for interaction becoming easier over time, visible in Table 3 and Figure 7, points to a short but meaningful adaptation process. Users were not interacting in a completely unconstrained conversational way, but they were learning, even within a brief activity, how to communicate more effectively with the robot. That is a useful finding for this type of awareness-oriented setup, because it suggests that short challenge-based experiences may already be enough to move participants beyond novelty and into a more informed understanding of how robot interaction actually works.

One of the more interesting patterns in the results is the separation between physical safety and operational confidence. Participants generally felt safe around the robot, but trust in its dependable behavior was lower than several of the other interaction measures. That distinction matters. It suggests that users may quickly become comfortable with the robot's presence, while still reserving judgment about its consistency, robustness, or reliability in action. In practical terms, making a robot appear approachable is not the same as making it feel dependable. For LLM-enabled systems in particular, that gap is especially important, because the flexibility of language-based interaction can make the system feel intuitive while still leaving room for ambiguity in execution.

The qualitative responses reinforce that reading of the results. Participants repeatedly pointed to voice-command interaction, the opportunity to engage directly with a humanoid robot, and the team-based challenge format as some of the strongest aspects of the experience, while concerns clustered around misinterpretation of commands, reliability or failure risk, and over-trust in automation. These themes, summarized in Table 4 and Figure 8, align well with the quantitative pattern rather than contradicting it. The challenge seems to have succeeded precisely because it exposed participants not only to the appeal of the technology, but also to its current friction points.

Taken together, these findings suggest that the Humanoid Challenge was more than a standalone demonstration. Beyond generating positive reactions, it provided a structured setting in which participants could directly experience both the opportunities and the limitations of interacting with an embodied AI system. By combining direct interaction, collaboration, time pressure, task progression, and post-event reflection, the challenge created an experience that was both engaging and informative. In this way, the activity supported not only interest in robotics, but also a more grounded understanding of what human-robot interaction may look like in practice.

More broadly, this has implications for how organizations may approach early exposure to embodied AI. In domains such as ports, logistics, construction, or industrial operations, adoption depends not only on technical capability, but also on how future users interpret the robot's usefulness, limitations, and role within existing workflows. Interactive formats such as the one presented here may therefore serve as a practical intermediate step between passive demonstrations and full operational deployment. They create a space where users can encounter the technology, test their assumptions, and begin identifying both opportunities and limitations in a relatively low-risk setting.

Overall, the findings suggest that LLM-enabled humanoid interaction can provide a promising foundation for robotics awareness activities in real organizational environments. At the same time, they also make clear that the long-term value of such activities will depend on continued improvements in command interpretation, responsiveness, predictability, and reliability. These are not peripheral technical details, but central factors in how users come to understand and evaluate the practical role of robots in collaborative settings.

## 6. Limitations and Future Work

Some limitations should be considered when interpreting the findings of this study. First, the evaluation relied primarily on self-reported survey responses, which capture participant perceptions but do not provide direct evidence of system performance. The study did not include objective interaction measures such as command success rates, response latency, navigation accuracy, or task completion efficiency. In addition, the respondents were drawn from a single organizational context, which limits the extent to which the findings can be generalized to other industrial settings, user populations, or task environments. Taken together, these factors mean that the present study should be understood as an initial field-based evaluation of the proposed method rather than as a controlled validation of LLM-enabled humanoid interaction in operational deployment conditions.

Future work should expand the evaluation in both technical and methodological directions. On the technical side, future studies should combine perception-based assessment with objective performance metrics, including command interpretation accuracy, execution success rate, response time, recovery from failed actions, and task completion outcomes. Interaction logs could also be used to analyze how users adapt their phrasing over time and which types of commands are more likely to produce ambiguity or failure. On the methodological side, the robotics awareness framework introduced in this paper should be tested across multiple organizations and participant profiles to examine its transferability beyond the present case. Future iterations could also compare direct participants and observers more explicitly in order to better understand the relative value of hands-on interaction versus structured observation. Finally, subsequent versions of the challenge may incorporate additional robotic platforms, including collaborative industrial robots, to explore how the method can be extended from humanoid-centered interaction toward broader multi-platform robotics awareness in industrial and operational environments.

## 7. Conclusion

This paper presented a challenge-based approach for robotics awareness built around an LLM-enabled humanoid robot activity conducted with employees of AD Ports Group. Through a logistics-inspired interaction scenario, participants engaged with a humanoid robot using natural-language voice commands, allowing the study to examine both the event experience and the direct interaction experience in an organizational setting.

The findings showed a positive overall reception of the activity, with high ratings for satisfaction, increased interest in robotics and artificial intelligence, and improved understanding of human-robot collaboration. Participants who interacted directly with the robot also reported positive perceptions of the interaction, while still pointing to important challenges related to command interpretation, predictability, responsiveness, and reliability. These results indicate that even brief, structured interaction can support meaningful familiarization with embodied AI, while simultaneously revealing the current limitations of LLM-enabled control in practice.

Positioned within the broader context of industrial robotics adoption, the results reinforce the importance of experiential exposure as a complement to technical deployment. While prior industry-led implementations and pilot programs have shown that workers gradually adapt to robotic systems through day-to-day interaction, such exposure is often unstructured and embedded within operational rollout processes. The approach presented in this paper contributes a more explicit and replicable framework, in which awareness is intentionally designed through a structured, task-driven activity that combines direct interaction, collaboration, and reflection within an operationally relevant scenario. In addition, this kind of low-stakes, hands-on format may help reduce initial hesitation and resistance by allowing potential users to engage with the system in a supportive setting before encountering it in a formal deployment context.

Overall, the study suggests that structured challenge-based activities can provide a useful and scalable method to expose non-specialist audiences to embodied AI in a direct and operationally meaningful way. By leveraging natural-language interaction as an accessibility layer, such approaches can lower barriers to engagement while preserving task realism and interaction complexity. As LLM-enabled robotic systems continue to evolve, combining interactive exposure with human-centered evaluation may help organizations not only assess technical capabilities, but also develop a more grounded understanding of how these systems may be integrated into real workflows. In this sense, robotics awareness activities of the kind presented here may serve as an important intermediate step between passive demonstrations and full-scale deployment, supporting more informed, reflective, and effective adoption of human–robot collaboration technologies.

## Acknowledgments

This work was partially supported by different Centers at NYUAD. In particular, the Center for Sand Hazards and Opportunities for Resilience, Energy, and Sustainability (SHORES), the Center for Interacting Urban Networks (CITIES), funded by Tamkeen under the NYUAD Research Institute Award CG001, and the Center for Artificial Intelligence and Robotics (CAIR). Part of this research benefited from the resources of the Core Technology Platform (CTP) at New York University Abu Dhabi (NYUAD), particularly the CTP’s KINESIS Lab. The authors would also like to acknowledge AD Ports Group and their Corporate Innovation Department for their collaboration and support in the organization and execution of the challenge that motivated this research.

## References

Ahn, M., Brohan, A., Brown, N., Chebotar, Y., Cortes, O., David, B., Finn, C., Fu, C., Gopalakrishnan, K., Hausman, K., Herzog, A., Ho, D., Hsu, J., Ibarz, J., Ichter, B., Irpan, A., Jang, E., Ruano, R. J., Jeffrey, K., … Zeng, A. (2022). Do As I Can, Not As I Say: Grounding Language in Robotic Affordances (Version 2). arXiv. https://doi.org/10.48550/ARXIV.2204.01691

Brohan, A., Brown, N., Carbajal, J., Chebotar, Y., Chen, X., Choromanski, K., Ding, T., Driess, D., Dubey, A., Finn, C., Florence, P., Fu, C., Arenas, M. G., Gopalakrishnan, K., Han, K., Hausman, K., Herzog, A., Hsu, J., Ichter, B., … Zitkovich, B. (2023). RT-2: Vision-Language-Action Models Transfer Web Knowledge to Robotic Control (Version 1). arXiv. https://doi.org/10.48550/ARXIV.2307.15818

Driess, D., Xia, F., Sajjadi, M. S. M., Lynch, C., Chowdhery, A., Ichter, B., Wahid, A., Tompson, J., Vuong, Q., Yu, T., Huang, W., Chebotar, Y., Sermanet, P., Duckworth, D., Levine, S., Vanhoucke, V., Hausman, K., Toussaint, M., Greff, K., … Florence, P. (2023). PaLM-E: An Embodied Multimodal Language Model (Version 1). arXiv. https://doi.org/10.48550/ARXIV.2303.03378

Gopee, M. A., Prieto, S. A., & García De Soto, B. (2025, June 17). Enhancing Human-Machine Interaction with On-Device Large Language Models in Construction Robotics: A Case Study on Safety Applications. Proceedings of the 14th Creative Construction Conference (CCC 2025). 14th Creative Construction Conference, Hotel Pinija, Petrcane, Zadar, Croatia. https://doi.org/10.22260/CCC2025/0041

Gopee, M. A., Prieto, S. A., & García De Soto, B. (2026). Self-hosted multimodal large language models for speech-driven perception and navigation in construction robotics. Automation in Construction, 183, 106805. https://doi.org/10.1016/j.autcon.2026.106805

Hancock, P. A., Billings, D. R., Schaefer, K. E., Chen, J. Y. C., De Visser, E. J., & Parasuraman, R. (2011). A Meta-Analysis of Factors Affecting Trust in Human-Robot Interaction. Human Factors: The Journal of the Human Factors and Ergonomics Society, 53(5), 517–527. https://doi.org/10.1177/0018720811417254

Huang, W., Wang, C., Zhang, R., Li, Y., Wu, J., & Fei-Fei, L. (2023). VoxPoser: Composable 3D Value Maps for Robotic Manipulation with Language Models (Version 2). arXiv. https://doi.org/10.48550/ARXIV.2307.05973

Kitano, H., Asada, M., Kuniyoshi, Y., Noda, I., Osawai, E., & Matsubara, H. (1998). RoboCup: A challenge problem for AI and robotics. In H. Kitano (Ed.), RoboCup-97: Robot Soccer World Cup I (Vol. 1395, pp. 1–19). Springer Berlin Heidelberg. https://doi.org/10.1007/3-540-64473-3_46

Krüger, J., Lien, T. K., & Verl, A. (2009). Cooperation of human and machines in assembly lines. CIRP Annals, 58(2), 628–646. https://doi.org/10.1016/j.cirp.2009.09.009

Liang, J., Huang, W., Xia, F., Xu, P., Hausman, K., Ichter, B., Florence, P., & Zeng, A. (2022). Code as Policies: Language Model Programs for Embodied Control (Version 4). arXiv. https://doi.org/10.48550/ARXIV.2209.07753

Parasuraman, R., & Riley, V. (1997). Humans and Automation: Use, Misuse, Disuse, Abuse. Human Factors: The Journal of the Human Factors and Ergonomics Society, 39(2), 230–253. https://doi.org/10.1518/001872097778543886

Pasparakis, A., De Vries, J., & De Koster, R. (2023). Assessing the impact of human–robot collaborative order picking systems on warehouse workers. International Journal of Production Research, 61(22), 7776–7790. https://doi.org/10.1080/00207543.2023.2183343

Pedersen, M. R., Nalpantidis, L., Andersen, R. S., Schou, C., Bøgh, S., Krüger, V., & Madsen, O. (2016). Robot skills for manufacturing: From concept to industrial deployment. Robotics and Computer-Integrated Manufacturing, 37, 282–291. https://doi.org/10.1016/j.rcim.2015.04.002

Prieto, S. A., Giakoumidis, N., & García De Soto, B. (2024). Multiagent robotic systems and exploration algorithms: Applications for data collection in construction sites. Journal of Field Robotics, 41(4), 1187–1203. https://doi.org/10.1002/rob.22316

Sony, M., & Naik, S. (2020). Industry 4.0 integration with socio-technical systems theory: A systematic review and proposed theoretical model. Technology in Society, 61, 101248. https://doi.org/10.1016/j.techsoc.2020.101248

Venkatesh, V., & Bala, H. (2008). Technology Acceptance Model 3 and a Research Agenda on Interventions. Decision Sciences, 39(2), 273–315. https://doi.org/10.1111/j.1540-5915.2008.00192.x

Wurhofer, D., Meneweger, T., Fuchsberger, V., & Tscheligi, M. (2015). Deploying Robots in a Production Environment: A Study on Temporal Transitions of Workers' Experiences. In J. Abascal, S. Barbosa, M. Fetter, T. Gross, P. Palanque, & M. Winckler (Eds.), Human-Computer Interaction – INTERACT 2015 (Vol. 9298, pp. 203–220). Springer International Publishing. https://doi.org/10.1007/978-3-319-22698-9_14

Yu, J., Chen, Q., Prieto Ayllon, S. A., & García de Soto, B. (2024, June 3). Human-Robot Partnership: An Overarching Consideration for Interaction and Collaboration. 41st International Symposium on Automation and Robotics in Construction, Lille, France. https://doi.org/10.22260/ISARC2024/0160